\documentclass{article}

\usepackage{arxiv}

\usepackage[utf8]{inputenc} 
\usepackage[T1]{fontenc}    
\usepackage{hyperref}       
\usepackage{url}            
\usepackage{booktabs}       
\usepackage{amsfonts}       
\usepackage{nicefrac}       
\usepackage{microtype}      
\usepackage{graphicx}
\usepackage[numbers]{natbib}
\usepackage{doi}
\usepackage{amsmath}
\usepackage{enumitem}
\usepackage[dvipsnames]{xcolor}
\usepackage{colortbl}
\usepackage{subcaption}
\usepackage[font=small]{caption}
\usepackage{algorithm}
\usepackage{algorithmic}
\usepackage{xspace}
\usepackage{cleveref}

\def\modelname{LACE\xspace}
\definecolor{Gray}{gray}{0.90}

\newcommand{\inc}[1]{\ensuremath{_{\text{\textcolor{PineGreen}{($+$#1)}}}}}
\newcommand{\dec}[1]{\ensuremath{_{\text{\textcolor{RedOrange}{($-$#1)}}}}}
\newcommand{\iinc}[1]{\ensuremath{_{\text{\textcolor{PineGreen}{($-$#1)}}}}}

\newcommand{\eg}{e.g.}

\title{\modelname: Latent Visual Representation for Cross-Embodiment Learning}

\author{
	Yoo Sung Jang$^1$ \quad Kanchana Ranasinghe$^2$ \quad Cristina Mata$^1$ \AND
	Yichi Zhang$^1$ \quad Jorge Mendez-Mendez$^1$ \quad Michael S. Ryoo$^1$ \\
	\\
	$^1$Stony Brook University \quad $^2$Salesforce AI Research
}
\date{}

\renewcommand{\shorttitle}{\modelname: Latent Visual Representation for Cross-Embodiment Learning}

\hypersetup{
pdftitle={\modelname: Latent Visual Representation for Cross-Embodiment Learning},
pdfsubject={cs.RO, cs.CV, cs.LG},
pdfauthor={Yoo Sung Jang, Kanchana Ranasinghe, Cristina Mata, Yichi Zhang, Jorge Mendez-Mendez, Michael S. Ryoo},
pdfkeywords={Cross-Embodiment Learning, Dexterous Manipulation, Robot Learning},
}

\begin{document}
\maketitle

\begin{figure}[h]
\centering
\includegraphics[width=0.90\linewidth]{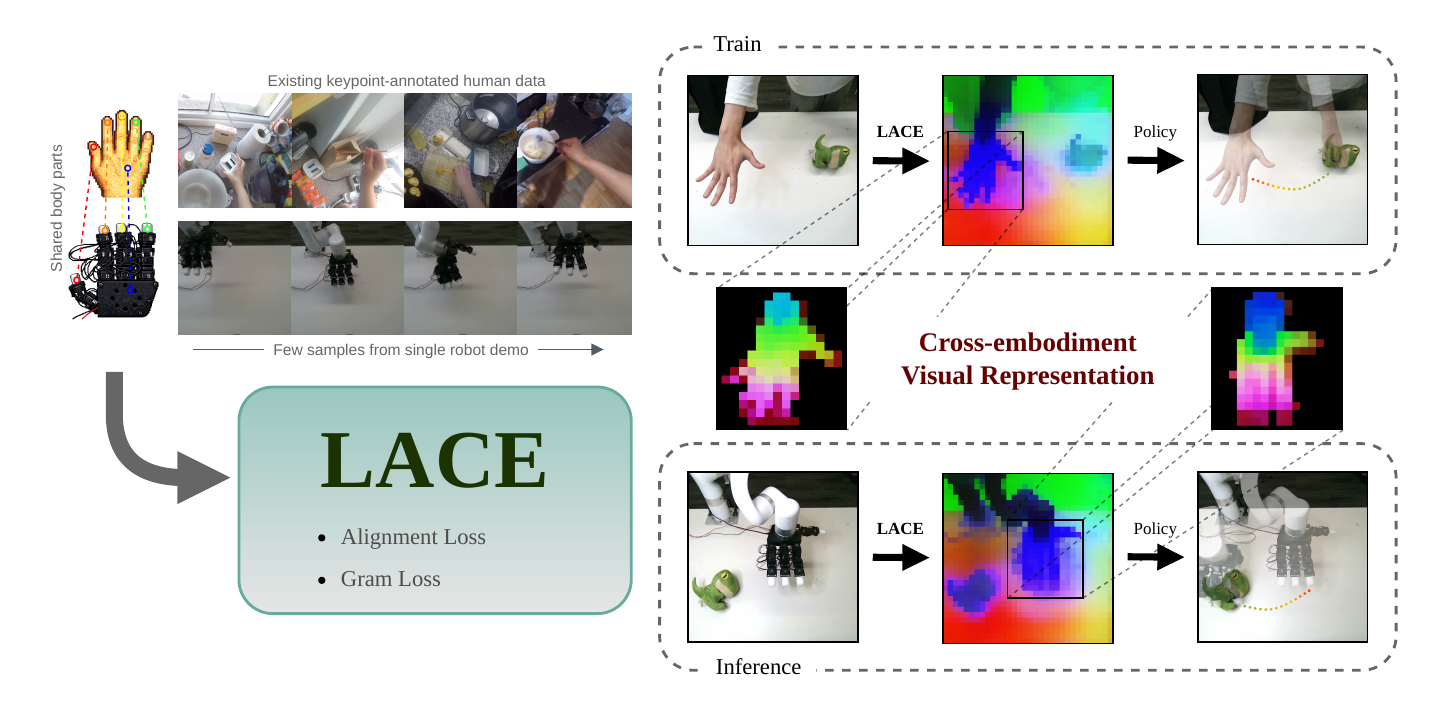}
\caption{We present \modelname{}, a framework for cross-embodiment visual representation alignment.
Across embodiments, they share semantically similar body parts. Leveraging this, \modelname minimizes the visual gap across embodiments in latent feature space.}
\label{fig:teaser}
\end{figure}

\begin{abstract}
Cross-embodiment learning from human demonstrations is hindered by the visual gap between human and robot embodiments.
While self-supervised learning (SSL) backbones encode rich inter-class semantics of general objects, we show they fail to establish correspondence between human and robot hands.
We propose \modelname, a framework that aligns human and robot visual representations in the latent space of these backbones by leveraging correspondences between shared body parts across embodiments as sparse supervision.
These annotations can be automatically obtained via forward kinematics, and single robot demonstration is sufficient to train the model.
Our semantic alignment loss matches distributions incurred by corresponding features, lifting patch-level supervision to semantic-level alignment, while a Gram loss preserves pretrained feature quality.
This alignment enables robot policies to leverage abundant human data when robot demonstrations are scarce: in zero-shot transfer, policies using \modelname-DINO outperform those using DINO by a large margin (65\%), 
with consistent gains in low-data regimes and out-of-distribution environments.
Our code will be released publicly.
\end{abstract}

\keywords{Cross-Embodiment Learning \and Dexterous Manipulation \and Robot Learning}

\section{Introduction}
\label{sec:intro}

Data scarcity is a long-standing challenge in robot learning.
Despite recent efforts to collect large-scale real world datasets~\cite{openx2024,droid2024,walke2023bridgedata},
such data remains limited to non-dexterous robot arms.
While there is growing interest in scaling datasets for dexterous hands, they have yet to reach comparable size or diversity.

Instead, many works aim to transfer behaviors from human hand demonstrations~\cite{qiu2025humanoidpolicyhuman,Ran25LangToMo},
as human video data is readily available on the internet.
However, human and robot hands are visually distinct, creating a bottleneck for smooth domain transfer.
A growing body of work addresses this visual gap across embodiments \cite{lepert2025shadow, chen2024miragecrossembodimentzeroshotpolicy, lepert2025phantom, lepert2025masquerade,Ran26FOFPred}, but these methods often focus on modifying representations in pixel space.
These pixel-space methods have several drawbacks: they discard embodiment-specific visual information \cite{lepert2025shadow,haldar2025point}, require additional image processing at inference time~\cite{gu2023rttrajectoryrobotictaskgeneralization,Ran26FOFPred,wen2023any}, or are limited to a specific target embodiment~\cite{chen2024miragecrossembodimentzeroshotpolicy, lepert2025phantomtrainingrobotsrobots, lepert2025masquerade}.

We propose to bridge the cross-embodiment visual gap in latent space.
Self-supervised learning (SSL) backbones have become the de facto visual representation in modern policies~\cite{kim2024openvla, black2026pi0visionlanguageactionflowmodel, brohan2022rt, zitkovich2023rt, qiu2025humanoidpolicyhuman, team2024octo}, which makes latent-space alignment a natural fit. 
Our approach bypasses the aforementioned limitations of pixel-space methods.
Moreover, it allows us to leverage the semantic structure that pretrained backbones already encode towards cross-embodiment alignment during training.

As a concrete instantiation of our method, we fine-tune DINOv3 (hereafter DINO) using an existing keypoint-annotated human dataset and a small set of robot images whose annotations are obtained automatically from forward kinematics.
We supervise alignment through correspondence between shared body parts across embodiments (e.g., human thumb to robot thumb).
Our method leverages the semantic priors that pretrained backbones already encode, both to lift this keypoint-level supervision to semantic-level alignment and to regularize training.
We term this framework \textbf{L}atent~\textbf{A}lignment~for~\textbf{C}ross-\textbf{E}mbodiment~(\modelname) and denote backbones fine-tuned under this paradigm by prefixing \modelname (e.g., \modelname-DINO); while we focus on DINO for its strength and popularity in robot learning, the approach readily transfers to other self-supervised backbones that encode semantic structure.

Our main contributions are:
\begin{itemize}[leftmargin=2em,noitemsep,topsep=0.0ex,itemsep=-1.0ex,partopsep=0ex,parsep=1ex]
    \item We identify that the visual gap between human and robot embodiments propagates to latent visual embeddings and propose a fine-tuning approach that leverages the backbone's inherent semantic priors to minimize this gap.
    \item We achieve cross-embodiment semantic alignment using frames sampled from a single robot demonstration - or even a single keypoint-annotated frame - while preserving the general semantic features, avoiding catastrophic forgetting.
    \item We empirically validate in real-world experiments that latent-space alignment benefits policy learning in low robot data regimes and enables generalization to unseen environments.
\end{itemize}

\section{Related Works}

\paragraph{Visual Gap in Cross-Embodiments Learning}
Robot learning across embodiments is challenged by the visual appearance gap between the source and target embodiment. 
One approach is to sidestep the visual gap using wrist-mounted cameras, which exclude the embodiment from visual observation~\cite{chi2024universal,Rayyan2025MVUMIAS,Cai2025InNOnSE,Xu2026HoMMILW,Kareer2024EgoMimicSI,Hoque2025EgoDexLD}.
However, wrist views omit crucial context such as visual robot state and holistic scene layout.

Several recent methods handle the visual gap by modifying images directly in pixel space~\cite{kareer2025egomimic, lepert2025shadow,chen2024miragecrossembodimentzeroshotpolicy,lepert2025phantomtrainingrobotsrobots,lepert2025masquerade,Fan2026RoboPaintFH}.
These approaches use techniques such as segmentation masks or inpainting to make observations from different embodiments appear visually similar.
However, pixel-space modifications have key drawbacks: they discard embodiment-specific information such as morphology, require additional image processing during deployment, and often the policy is committed to a specific target embodiment.
Other work provides embodiment-agnostic visual cues to guide the policy~\cite{gu2023rttrajectoryrobotictaskgeneralization, wen2023any,Ran25LangToMo,Nguyen2025PixelMD,Hu2024VideoPP,Goswami2025WorldMC}, but still requires additional compute during inference.

An alternative direction operates in latent space: recent work observes that shared representations can emerge through co-training on mixed human-robot data~\cite{kareer2025emergencehumanrobottransfer,punamiya2025egobridge,Luo2026BeingH05SH,Kim2025UniSkillIH,Xu2023XSkillCE}.
However, these methods require large and diverse datasets, or temporally aligned demonstrations across embodiments which need to be carefully collected.

Our method does not require a temporally or spatially aligned dataset nor a large dataset, yet we acheive the latent representation alignment leveraging the encoded semantics of a pretrained SSL backbone.
We fine-tune the backbone using samples from as few as one demonstration, achieving cross-embodiment alignment without large datasets or inference overhead.

\paragraph{Self-Supervised Visual Backbones}

Self-supervised models such as DINO-v2~\&~v3, SigLIP and V-JEPA2~\cite{oquab2023dinov2, simeoni2025dinov3, tschannen2025siglip, assran2025v} are widely adopted as visual backbones in modern vision-language-action models~\cite{kim2024openvla, shang2024theia, di2024dinobot, qiu2025humanoidpolicyhuman, xie2025human2robot}.
Trained on internet-scale data, these models produce dense, locality-aware, and semantically rich features, which have proven useful for downstream tasks including policy learning in robotics.

However, these properties do not fully transfer to cross-embodiment learning settings in robotics, especially with dexterous manipulators \cite{Yuan2024CrossEmbodimentDG,Wu2026UniMorphGraspDM,Wu2025CEDexCD,Zhang2025MorphArtGraspMC,He2025ScalingCW,Fay2025CrossembodiedCF,Fang2025AnyDexGraspGD,Zhang2025RobustDexGraspRD,Ga2025CrossingTH,Bi2025HRDTHM}.
Self-supervised representations are typically pretrained on internet images that contain less robotic embodiments or manipulation interactions.
As we show in Section~\ref{sec:motivation}, pretrained DINO (self-supervised) features fail to establish reliable semantic correspondence between human and robot hands despite their functional and structural similarities.
While robotics-specific self-supervised encoders exist~\cite{nair2022r3m,radosavovic2023real,majumdar2023we,hou20254d}, their ability to quickly adapt to embodiments unseen during pretraining remains underexplored.
Our work addresses this gap by explicitly aligning features across embodiments using only few samples, while preserving the rich semantic properties of the pretrained SSL backbone.

\begin{table}[t]
    \centering
    \caption{\textbf{Keypoint alignment metrics comparison.} We evaluate keypoint alignment on different embodiment pair settings. DINO struggles with human–robot correspondence; \modelname~addresses this gap. This improvement generalizes beyond DINO to other SSL backbones (Appendix~\ref{sec:other_ssl}).}
    \vspace{0.5em}
    \label{tab:pretrained_alignment}
    \def\arraystretch{1.1}  
    \setlength\tabcolsep{1.0em}  
    \scalebox{0.88}{
    \begin{tabular}{lcccc}
        \toprule
        & \multicolumn{2}{c}{EPE $\downarrow$} & \multicolumn{2}{c}{Cos $\uparrow$} \\
        \cmidrule(lr){2-3} \cmidrule(lr){4-5}
        & DINO & Ours & DINO & Ours \\
        \midrule \rowcolor{Gray}
        \textbf{Human--Robot} & \textbf{87.04} & \textbf{19.36} & \textbf{0.473} & \textbf{0.910} \\
        Human--Human & 34.94 & 17.57 & 0.679 & 0.903 \\
        Robot--Robot & 23.00 & 16.00 & 0.824 & 0.948 \\
        \bottomrule
    \end{tabular}
    }
\end{table}

\begin{figure}[t]
\centering
\includegraphics[width=0.8\linewidth]{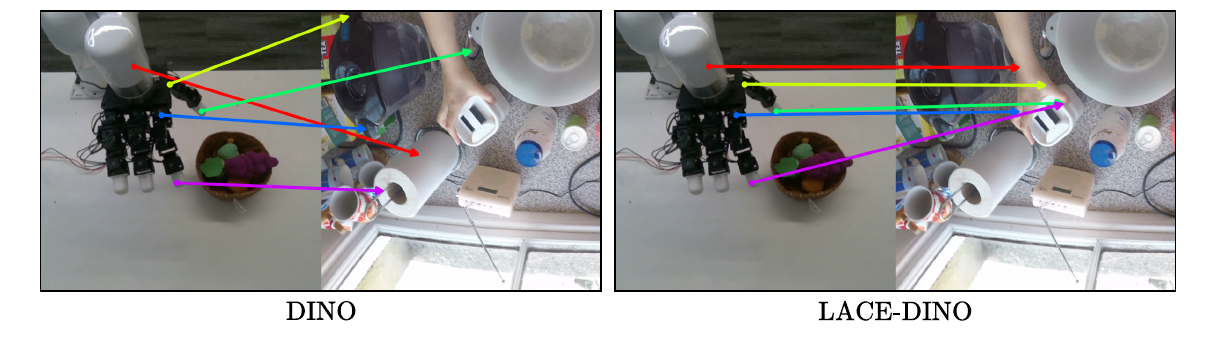}
\caption{
\textbf{Cross-embodiment correspondence gap.} 
Cross-embodiment (H-R) correspondence is weak in DINO; \modelname~achieves strong correspondence.}
\label{fig:motivation}
\end{figure}
\section{Motivation}
\label{sec:motivation}

Semantically rich DINO features provide semantic correspondence not only between image pairs of the same class
but also across different domains (e.g., see Fig 10 in~\cite{oquab2023dinov2}).
We investigate whether this property extends to the cross-embodiment setting (e.g. human to robot) in robotic tasks.

We collect 100 keypoint-annotated human hand images from EpicKitchen~\cite{darkhalil2022epic} and 100 robot hand images from a real-world lab setting.
We evaluate correspondence across all image pairs in three settings: human--human (H-H), robot--robot (R-R), and human--robot (H-R).
We evaluate correspondence using two metrics: \textit{End Point Error (EPE)}~\cite{zb2017hand, toshev2014deeppose}, which finds the most similar patch in the target image via cosine similarity of DINO features and measures the pixel distance to the ground truth keypoint; and \textit{Cosine Similarity (Cos)}, which directly measures feature similarity between ground truth keypoint patches across paired images.

Table~\ref{tab:pretrained_alignment} shows that pretrained DINO achieves reasonable within-domain correspondence (H-H, R-R) but fails on cross-embodiment pairs—H-R error is approximately 3$\times$ worse and feature similarity drops substantially.
Figure~\ref{fig:motivation} shows example correspondences from pretrained DINO, illustrating the misalignment in the H-R case.
Similar behavior is observed for other SSL models; see Appendix~\ref{sec:other_ssl} for details.

We posit that this failure is due to the domain gap in the pretraining dataset.
DINO was trained on LVD-1689M, a proprietary dataset collected from Instagram,
which lacks robotics data and in particular diverse robot embodiments.
Consequently, while DINO's pretrained features capture rich semantics for natural images,
they do not learn correspondence between human and robot hands.
This limitation likely extends to many computer vision SSL backbones~\cite{oquab2023dinov2,tschannen2025siglip,balestriero2025lejepa}.
Although some backbones are pretrained specifically for robotics~\cite{nair2022r3m,radosavovic2023real,majumdar2023we,hou20254d}, they may still struggle
to adapt to new robot embodiments unseen during pretraining—a common scenario given
the rapid evolution of robot hardware.

This absence of cross-embodiment alignment in the pretrained backbone motivates our fine-tuning approach.
We leverage the structural correspondence between human and robot hands—shared body parts—to establish alignment while preserving the backbone's semantic features.

\section{Method}
Embodiments within the same category share similar morphology and kinematics.
In particular, body parts of a dexterous robot hand can be mapped one-to-one to those of a human hand.
This structural correspondence provides a natural prior for aligning visual representations across embodiments.

\subsection{Problem Setup}

\noindent\textbf{Cross-Embodiment Imitation Learning.}
We study cross-embodiment imitation learning between humans and robots that share a common action space $\mathcal{A}$ but differ in visual observation spaces $\mathcal{O}_H$ and $\mathcal{O}_R$ due to appearance differences.
We consider a setting with abundant human demonstrations $\mathcal{D}_H = \{(o^H_t, a_t)\}$ but limited robot data $\mathcal{D}_R = \{(o^R_t, a_t)\}$, where the goal is to train a policy that transfers from human to robot and generalizes to unseen environments.
\smallskip

\noindent\textbf{Visual Latent Alignment.}
One key challenge is the visual domain gap: a policy trained on human observations $o^H$ cannot directly generalize to robot observations $o^R$ due to the embodiment difference.
Our goal is to fine-tune a pretrained visual encoder $f$ such that $f(o^H) \approx f(o^R)$ for semantically corresponding observations, mapping both embodiments to a shared latent state $s$.

We parameterize the policy as $\pi = g \circ f$, where $f$ is a visual encoder and $g$ is a policy network.
The encoder maps an observation $o$ to a latent state $s = f(o)$, and the policy outputs an action primitive $a$.
\smallskip

\noindent\textbf{Design Choice.}
We condition the policy solely on visual features, excluding proprioceptive input.
This design choice isolates the visual domain gap from other axes of cross-embodiment generalization (e.g., proprioceptive alignment), allowing us to directly evaluate contribution of visual representation alignment.

\subsection{Keypoint Pair Dataset}
We use keypoint correspondences across embodiments as supervision for alignment.
Let $\mathcal{E} = \{e_1, e_2, \ldots, e_n\}$ denote a set of embodiments (\eg, human hand, robot hand).
For each embodiment $e$, we have a dataset of images $\mathcal{I}_e$. Each image $I \in \mathcal{I}_e$ has associated keypoint annotations $\kappa(I) \in \mathbb{R}^{K \times 2}$ and visibility flags $v(I) \in \{0, 1\}^K$, where $K$ is the number of semantically corresponding parts shared across embodiments.

Training pairs are sampled from all embodiment combinations:
\[
\mathcal{P} = \{(I_i, I_j) : I_i \in \mathcal{I}_{e_a}, I_j \in \mathcal{I}_{e_b}, \; e_a, e_b \in \mathcal{E}\},
\]
including both cross-embodiment pairs ($e_a \neq e_b$) and intra-embodiment pairs ($e_a = e_b$).
For each pair, we compute the set of shared visible keypoints $\mathcal{K}_{ij} = \{k : v_i^k = 1 \land v_j^k = 1\}$ and apply the semantic alignment loss (Equation~\ref{eqn:alignment_loss}) over these keypoints.

This formulation requires no temporal synchronization or pose matching between embodiments—only semantic correspondence of body parts.
Robot keypoints are obtained automatically via forward kinematics; human keypoints are sourced from existing annotated datasets~\cite{pavlakos2024reconstructing} (Section~\ref{sec:implementation}).
We choose a human hand dataset with diverse scenes, backgrounds and objects.
Non-hand regions in these images implicitly serve as negative samples in our alignment loss, preventing collapse of non-embodiment features such as objects and backgrounds.

\subsection{\modelname : Latent Alignment for Cross-Embodiment}

Most SSL models' dense features exhibit strong semantic information, enabling robust correspondence even across different object classes.
However, as shown in Section~\ref{sec:motivation}, this property does not extend to human--robot correspondence.

We address this by fine-tuning a pretrained SSL model using keypoint supervision from a single demonstration.
Our training objective consists of two components: an alignment loss that encourages corresponding human and robot semantics to produce similar features, and a Gram loss that preserves the semantic structure of the original SSL features.

\begin{figure}[t]
\centering
\includegraphics[width=1.0\linewidth]
{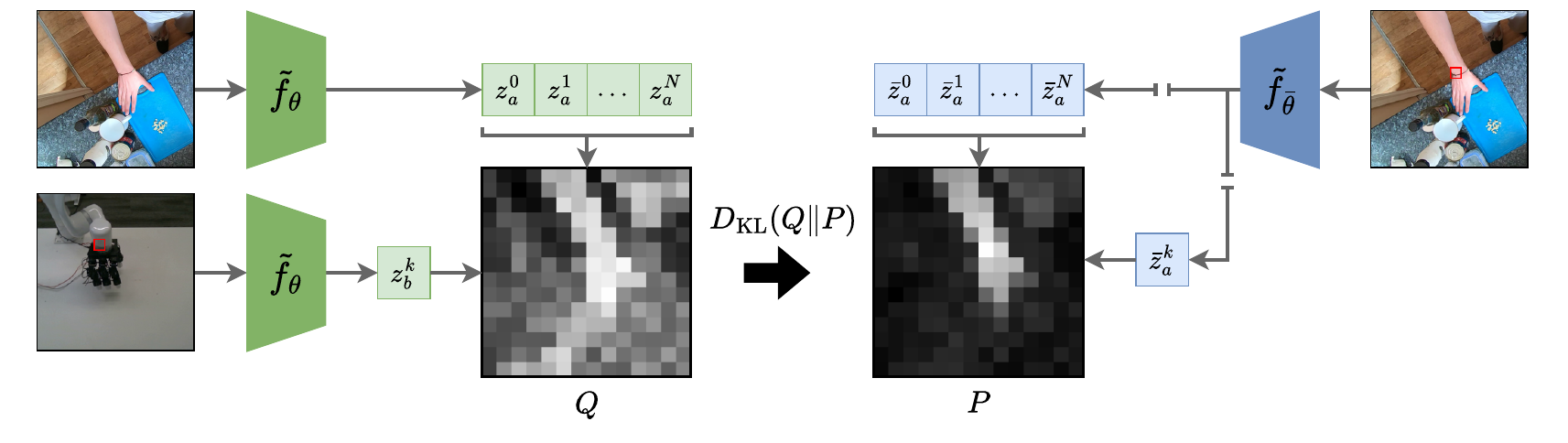}
\caption{\textbf{Semantic alignment loss visualization.} 
For a shared keypoint $k$ (denoted by red box), we match cross-similarity distribution $Q$ to self-similarity distribution $P$ by minimizing reverse KL divergence. 
}
\label{fig:alignment_loss}
\end{figure}

\subsubsection{Semantic Alignment Loss}
\label{sec:semantic_alignment_loss}

We sample image pairs across and within embodiments and identify visible keypoints in common.
Rich semantic structure emerges in SSL features through patch-wise relationships within an image.
Leveraging this, we align the semantic structures induced by corresponding keypoints (Figure~\ref{fig:alignment_loss}). 
\par Formally, let $f_\theta$ denote the trainable encoder and $f_{\bar{\theta}}$ denote its exponential moving average (EMA).
Using an EMA encoder provides a slowly evolving reference, which stabilizes training when the semantic structure differs substantially between embodiments.
We define L2-normalized encoders $\tilde{f} = f / \|f\|_2$.
Given an image pair $(I_a, I_b)$, we extract normalized patch features $Z = \tilde{f}_\theta(I)$, with $z^k$ denoting the normalized patch feature at shared keypoint $k$.
We compute:
\begin{align*}
P &= \mathrm{softmax}(\bar{z}_a^{k\top} \bar{Z}_a / \tau), \\
Q &= \mathrm{softmax}(z_b^{k\top} Z_a / \tau).
\end{align*}

$P$ captures the target semantic structure induced by a keypoint, while $Q$ captures the proposed semantic structure induced by its corresponding keypoint from another embodiment.
The alignment loss uses reverse KL divergence:
\begin{equation}
\label{eqn:alignment_loss}
\mathcal{L}_{\text{align}} = D_{\mathrm{KL}}(Q \| P).
\end{equation}

This loss encourages corresponding keypoint patches to induce similar semantic structure over the image, \textit{thereby lifting keypoint-level supervision to semantic-level supervision across embodiments.}

\subsubsection{Gram Loss}

Fine-tuning on a small, single-domain dataset risks overfitting and losing the semantic richness of DINO features.
Inspired by~\cite{simeoni2025dinov3}, we regularize the model by preserving the pairwise feature similarities between the trainable encoder and the EMA encoder.

For each image $I$, let $Z = \tilde{f}_\theta(I)$ and $\bar{Z} = \tilde{f}_{\bar{\theta}}(I)$ denote patch features from the trainable and EMA encoders, respectively.
The Gram loss penalizes deviation from the EMA encoder's Gram matrix:
\begin{equation}
\label{eqn:gram_loss}
\mathcal{L}_{\text{gram}} = \mathrm{MSE}(Z Z^\top, \bar{Z} \bar{Z}^\top).
\end{equation}
This encourages the fine-tuned model to retain the relative structure among patch features, preventing aggressive drift while allowing cross-embodiment alignment.
We show in Sections~\ref{sec:gram_ablation} and Appendix~\ref{sec:feature_drift} that the Gram loss is crucial for preserving feature quality and distribution. 

\begin{figure}[t]
    \centering
    \includegraphics[width=0.8\linewidth]{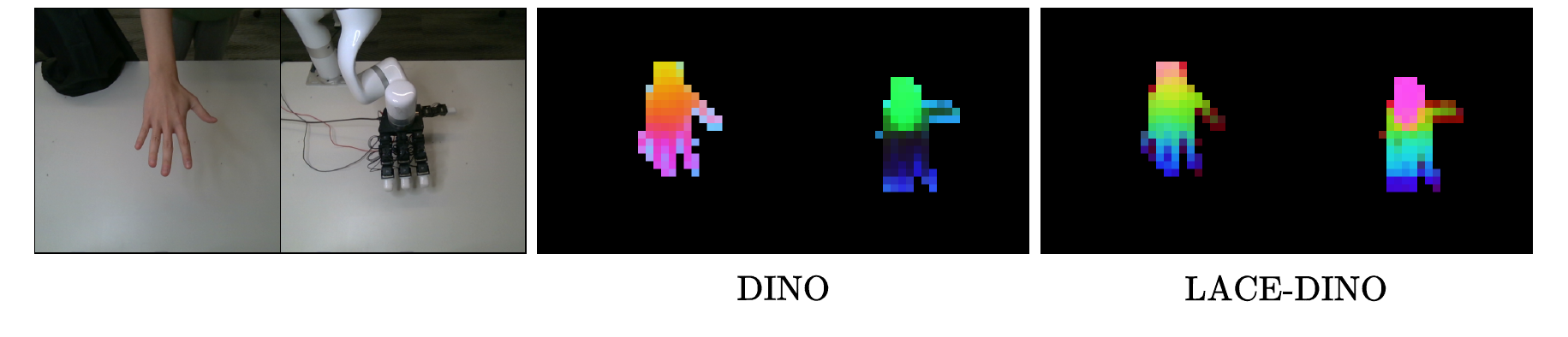}
    \caption{
    \textbf{Cross-embodiment feature alignment comparison.}
    PCA is computed jointly on human and robot hand features.
    Matching colors indicate semantically corresponding regions across embodiments.
    }
    \label{fig:masked_pca}
\end{figure}

Our total training objective combines both losses:
\begin{equation}
\label{eqn:total_loss}
\mathcal{L} = \mathcal{L}_{\text{align}} + \lambda \mathcal{L}_{\text{gram}}.
\end{equation}
where $\lambda$ controls the strength of regularization.
With this objective, we achieve cross-embodiment semantic alignment as shown in Figure~\ref{fig:masked_pca}.

\section{Experiments}

\subsection{Implementation Detail of \modelname-DINO}
\label{sec:implementation}

\paragraph{Keypoint Dataset}
\label{sec:keypoint_dataset}

For human hand images, we use the EpicKitchen subset of the HInt dataset~\cite{pavlakos2024reconstructing, darkhalil2022epic}, which provides egocentric view images with 21 keypoint annotations and occlusion flags.
We filter for frames where only the right hand is visible using~\cite{wu2019detectron2, potamias2025wilor}.

For robot hand images, we use an xArm with a LeapHand~\cite{shaw2023leaphand} end-effector.
We collect a single demonstration via Apple Vision Pro (AVP) teleoperation~\cite{park2024avp} and sample 100 frames.
Robot keypoints are computed from forward kinematics given known camera parameters, with visibility determined by renderer segmentation masks.
LeapHand has four fingers, so we drop annotations for the little finger; alternatively, one could map it to the ring finger.
We construct training pairs using shared visible keypoints, including both cross-embodiment (human--robot) and intra-embodiment (human--human, robot--robot) pairs.
We apply random crop-resize and Gaussian blur to both domains, augmenting the dataset to 1K images per embodiment.
\paragraph{Training}
We fine-tune a pretrained DINOv3 ViT-S backbone using AdamW with a learning rate of $1 \times 10^{-4}$ for 10K iterations.
Input images are resized to $224 \times 224$.
We adopt a student-teacher setup with EMA decay of 0.999, Gram loss weight $\lambda = 1.0$, and temperatures $\tau = 1.0$.
These settings are used throughout unless otherwise stated.

\subsection{Representation Quality for Cross-Embodiment Transfer}
\label{sec:rep_quality}

Following standard practice~\cite{oquab2023dinov2, simeoni2025dinov3}, we train a simple task-specific head for each downstream task.
In particular, we train attentive linear probes on two downstream tasks---pose estimation and localization---to evaluate \modelname-DINO as a representation for cross-embodiment transfer.
The key setup is \textit{zero-shot embodiment transfer}: probes are trained on human data only and evaluated on robot data.
\smallskip

\noindent\textbf{Dataset.}
\label{sec:data_pipeline}
For the downstream tasks, we collect 100 human demonstrations for training and 10 robot demonstrations for testing, sampling ${\sim}50$ images each with end-effector and target object labels.
End-effector positions are obtained via AVP (human) or FK (robot), projected to pixel space, and processed differently for each downstream task.
For the localization downstream task, we also process target object positions using OWL-ViT~\cite{minderer2022simple}, an open-vocabulary detection model queried with object descriptions (e.g., ``a green dinosaur doll''); we extract the bounding-box centers.
To test the generalization capability of the learned representation across different environments, we use a segmentation model~\cite{carion2025sam} to replace the lab backgrounds with kitchen and office scenes.
We train three separate \modelname-DINO checkpoints (lab, kitchen, office) and evaluate both downstream tasks in the lab setting.
\smallskip

\noindent\textbf{Training.}
We add sinusoidal position embeddings to the input features of linear attention probes.
All probes are trained for 10K iterations using AdamW with learning rate $1 \times 10^{-3}$ and evaluated on a held-out test set.

\begin{table}[t]
    \centering
    \caption{\textbf{Zero-shot transfer from human to robot.} Attentive linear probes trained on human data, evaluated on robot data. EPE in pixels (lower is better); all other metrics in \% (higher is better). Evaluation in identical environment for all methods. 
    }
    \label{tab:representation_eval}
    \vspace{0.5em}
    \def\arraystretch{1.1}  
    \setlength\tabcolsep{0.4em}  
    \scalebox{0.80}{
    \begin{tabular}{lllllll}
        \toprule
        & \multicolumn{3}{c}{Pose Estimation} & \multicolumn{3}{c}{Localization} \\
        \cmidrule(lr){2-4} \cmidrule(lr){5-7}
        & EPE $\downarrow$ & PCK@0.2 $\uparrow$ & AUC@0.5 $\uparrow$ & Emb $\uparrow$ & Obj $\uparrow$ & Both $\uparrow$ \\
        \midrule
        DINO & 57.5 & 28.6 & 48.7 & 3.2 & 57.9 & 1.6 \\
        \midrule
\modelname-DINO & 35.6 \iinc{21.9} & 72.4 \inc{43.8} & 68.2 \inc{19.5} & 40.5 \inc{37.3} & 65.8 \inc{7.9} & 23.2 \inc{21.6} \\
\modelname-DINO (Kitchen) & 24.8 \iinc{32.7} & 93.2 \inc{64.6} & 77.8 \inc{29.1} & 49.5 \inc{46.3} & 40.5 \dec{17.4} & 23.7 \inc{22.1} \\
\modelname-DINO (Office) & 34.4 \iinc{23.1} & 77.7 \inc{49.1} & 69.3 \inc{20.6} & 53.2 \inc{50.0} & 45.3 \dec{12.6} & 21.1 \inc{19.5} \\

        \bottomrule
    \end{tabular}
    }
\end{table}

\begin{figure}[t]
\centering
\includegraphics[width=\linewidth]{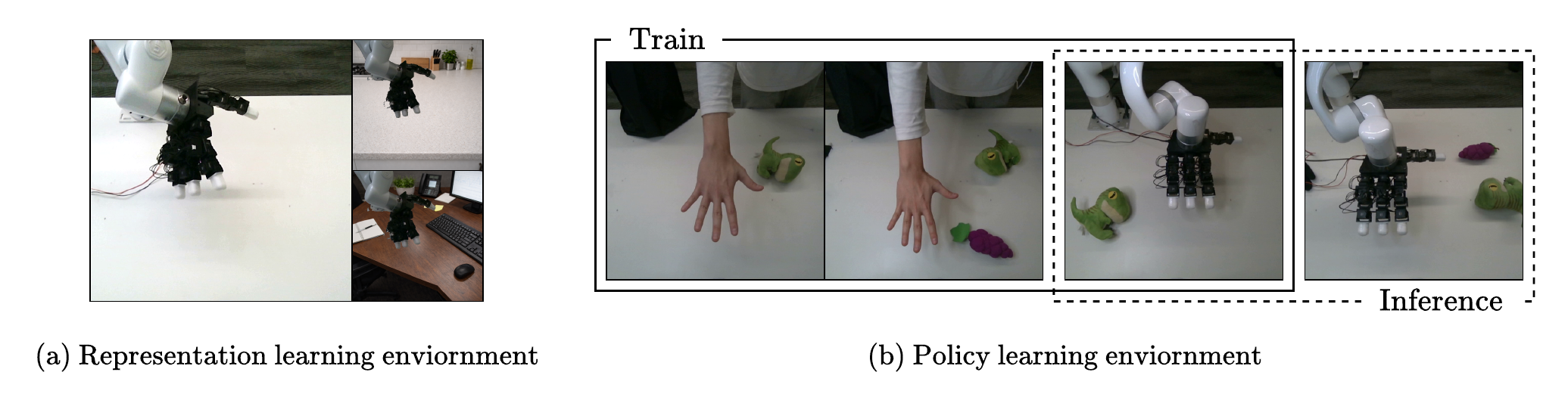}
\caption{\textbf{Real-world environment setup.} a) Lab (left), kitchen (right-up) and office (right-down) scenes for the representation learning. b) For policy learning envs from left to right: human in-domain, human OOD-env, robot in-domain, robot OOD-env. Both human and robot OOD-envs include a distractor object.}
\label{fig:real_world}
\end{figure}

\vspace{1em}
\paragraph{Pose Estimation}

Recovering robot state from visual observation is essential for policy learning.
Pose estimation, which captures the spatial configuration of the embodiment, tests whether \modelname can enable this in a zero-shot setting.

We predict the wrist and four fingertip positions in pixel coordinates.
The five keypoints from the data pipeline serve directly as regression targets.
Following standard evaluation protocols~\cite{zb2017hand, toshev2014deeppose}, we report End Point Error (EPE), Percentage of Correct Keypoints (PCK), and Area Under Curve (AUC).

\modelname-DINO substantially outperforms DINO across all metrics (Table~\ref{tab:representation_eval}), confirming that the learned representation successfully aligns body parts across embodiments.

\vspace{1em}
\paragraph{Localization}

While pose estimation tests body-part alignment, localization tests whether skills learned from human demonstrations generalize to robot observations.
In behavior cloning, a policy must understand where the embodiment is relative to the target object.

We predict the centers of bounding boxes for both the embodiment and target object.
For the embodiment, we construct a bounding box from the wrist and fingertip positions and use its center as the label; for the target object, we use the bounding-box center predicted from OWL-ViT.
We report hit rate: whether the predicted point falls within the ground truth bounding box.
We evaluate three cases: embodiment only (Emb), target object only (Obj), and both (Both).

The Emb metric is analogous to pose estimation---it measures embodiment localization.
The Obj metric is the key addition: it tests whether the model can locate the target object when observing an embodiment it has never seen during training.
This demonstrates how cross-embodiment semantic alignment can benefit skill transfer.

\modelname-DINO outperforms DINO on all metrics (Table~\ref{tab:representation_eval}).
The large improvement on Emb is expected given the pose estimation results.
For Obj, we observe that skills learned from human demonstrations transfer to robot observations. This effect is more apparent in Appendix~\ref{app:localization}, where we provide a         
controlled evaluation for a policy. 

\vspace{1em}
\paragraph{Environment Generalization}

We test the generalization of cross-embodiment alignment by training \modelname-DINO on images with kitchen \& office backgrounds, then evaluating downstream tasks in the same lab setting.
The environment-shifted variants perform comparably on pose estimation and embodiment localization (Table~\ref{tab:representation_eval}), indicating that \modelname is scene-independent.

\subsection{Cross-Embodiment Policy}
\label{sec:policy}
We evaluate whether \modelname-DINO enables policies to leverage human demonstrations for improved generalization.

\begin{table}[t]
\centering
\small
\setlength{\tabcolsep}{8pt}
\caption{\textbf{Success rate for object grasping.} We report success percentage (\%) with \texttt{successes/trials} within parentheses.}
\vspace{0.5em}
\label{tab:policy_eval_new}
\def\arraystretch{1.1}  
\setlength\tabcolsep{0.5em}  
\scalebox{1.0}{
\begin{tabular}{llcccc}
\toprule
& & \multicolumn{2}{c}{In-domain} & \multicolumn{2}{c}{OOD-env} \\
\cmidrule(lr){3-4} \cmidrule(lr){5-6}
Task & Regime & DINO & Ours & DINO & Ours \\
\midrule
Pick up dino & zero & 5  (1/20) & 60\inc{55} (12/20) &  0 (0/20) & 65\inc{65} (13/20) \\
             & few  & 10 (2/20) & 50\inc{40} (10/20) & 10 (2/20) & 60\inc{50} (12/20) \\
             & co   & 20 (4/20) & 65\inc{45} (13/20) & 40 (8/20) & 65\inc{25} (13/20) \\

\bottomrule
\end{tabular}
}
\end{table}

\begin{figure}[t]
\centering
\includegraphics[width=0.95\linewidth]{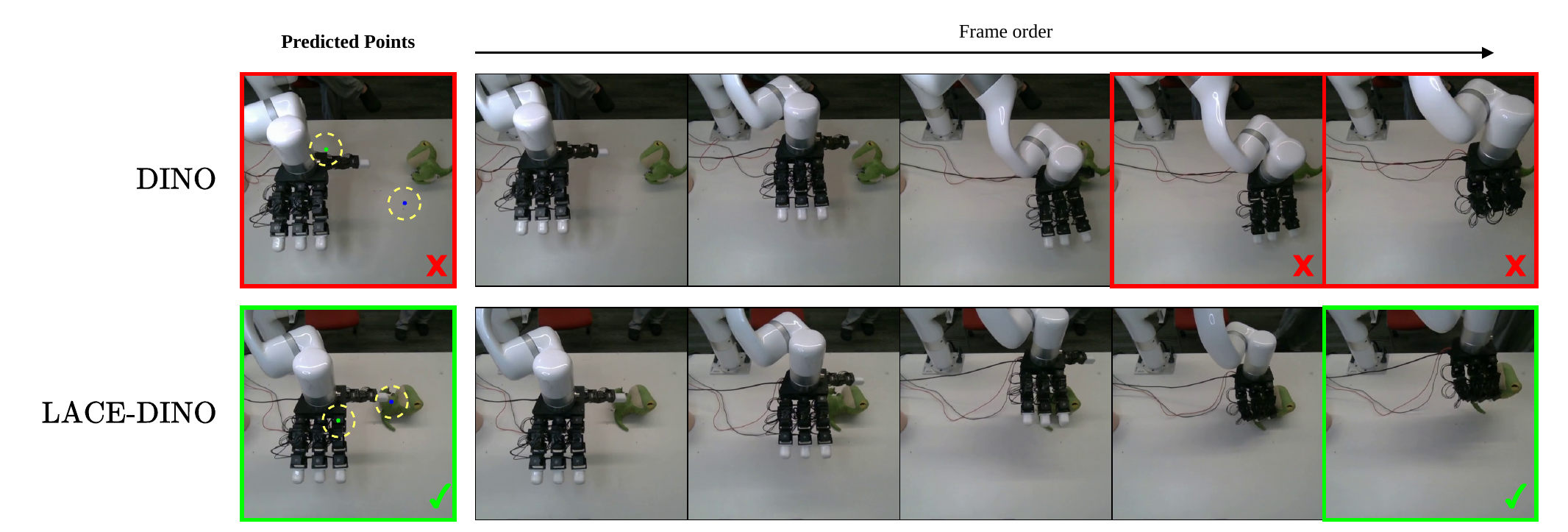}
\caption{\textbf{Real-world rollout examples.} On the ``Pick up dino'' task, the DINO-based policy locates the wrong target, while the \modelname-based policy predicts correctly. Predicted source and target points are shown in the first column.}
\label{fig:rollout}
\end{figure}

\noindent\textbf{Task }
We evaluate on an object grasping task where the policy must recover the robot state by localizing its end-effector and localize the target object from visual observations, 
which together enable grasping via motion primitives.

\noindent\textbf{Policy }
We modify DETR~\cite{carion2020end} as a policy by replacing the vision encoder with frozen \modelname-DINO; the baseline uses DINO. 
Additionally, the prediction head outputs two 2D points in pixel coordinates: end-effector position and target object position. 
The DETR architecture is initialized from pretrained weights to leverage object localization priors, with separate learning rates for pretrained parameters and the prediction head.

\noindent\textbf{Dataset.}
We collect a separate dataset for policy evaluation.
Human demonstrations are collected in two settings: distractor-free and with a distractor object.
Robot demonstrations are collected only in the distractor-free setting.
We use a total of 200 human and 50 robot demonstrations, sampling ${\sim}50$ images per demonstration.
Labels follow the same pipeline as Section~\ref{sec:data_pipeline}.

\noindent\textbf{Evaluation }
We report success rate for picking up the target object. Failure occurs when incorrect end-effector localization causes unsafe robot motion, triggering a safety stop, or when the policy fails to grasp the target object.
While success rate captures overall task performance, failures can arise from both localization errors and motion primitive variability. To disentangle these factors, we also provide a controlled localization analysis in Appendix~\ref{app:localization}.

We denote the distractor-free setting as \textit{in-domain} and the distractor setting as \textit{OOD-env} (Fig.~\ref{fig:real_world}). For training, we use all human demonstrations combined with varying amounts of robot data.
With this setup, we examine whether aligned representations can compensate for limited robot data, and whether they enable generalization to environments the robot has never encountered.

\paragraph{Data Regime}
Robot demonstrations are costly to collect, while human demonstrations are relatively cheap.
We test whether \modelname-DINO can leverage human data to reduce the need for robot demonstrations.
We evaluate across three data regimes based on the percentage of robot data used: 0\% (\textit{zero}), 10\% (\textit{few}), and 100\% (\textit{co}).
Table~\ref{tab:policy_eval_new} reports success rates across data regimes.
\modelname-DINO consistently outperforms DINO across all settings.
The gap is largest in the zero-shot regime, where \modelname-DINO achieves 60\% compared to 5\% for DINO in-domain.
\paragraph{Environment Generalization}
In the OOD-env setting, a distractor object is present that the robot has never seen—only human demonstrations include this scenario.
\modelname-DINO achieves 65\% in the zero-shot regime, compared to 0\% for DINO, demonstrating that aligned representations enable the robot to leverage human experience from unseen environments.
\vspace{0.5em}

These results confirm that latent-space alignment enables effective cross-embodiment transfer, allowing robots to benefit from human demonstrations even with limited robot data.

\begin{table}[t]
    \centering
    \small
    \caption{\textbf{Data efficiency of semantic alignment.} Human-to-robot correspondence metrics across varying numbers of training images. Even with a single demonstration, \mbox{\modelname-DINO} significantly outperforms the pretrained DINO baseline.}
    \vspace{0.5em}
    \label{tab:data_efficiency}
    \def\arraystretch{1.1}  
    \setlength\tabcolsep{1.0em}  
    \scalebox{0.9}{
    \begin{tabular}{lcccccc}
        \toprule
        & \multicolumn{2}{c}{H-R} & \multicolumn{2}{c}{H-H} & \multicolumn{2}{c}{R-R} \\
        \cmidrule(lr){2-3} \cmidrule(lr){4-5} \cmidrule(lr){6-7}
        & EPE $\downarrow$ & Cos $\uparrow$ & EPE $\downarrow$ & Cos $\uparrow$ & EPE $\downarrow$ & Cos $\uparrow$ \\
        \midrule
        DINO   & 87.04 & 0.473 & 34.94 & 0.679 & 23.00 & 0.824 \\
        \midrule
        \modelname-DINO \\
        \quad 1 img   & 23.81 & 0.861 & 17.58 & 0.906 & 24.20 & 0.901 \\
        \quad 10 img  & 19.90 & 0.903 & 17.84 & 0.905 & 18.30 & 0.936 \\
        \quad 100 img & 19.57 & 0.910 & 17.78 & 0.903 & 16.04 & 0.948 \\
        \bottomrule
    \end{tabular}
    }
\end{table}

\subsection{Data Efficiency of Learning Representation}
\label{sec:data_efficiency}

Extracting dense keypoint correspondences via forward kinematics requires careful engineering to map all matching body parts between embodiments.
As a simpler alternative, users can manually annotate visible keypoints on a small number of frames and still acquire cross embodiment alignment.
We ablate the number of training images to test whether \modelname-DINO still achieves strong alignment under this simplified setup.

We sample 1, 10, and 100 images from a single demonstration, augmenting each set to 1k training samples for fair comparison.
Using the same evaluation protocol as Section~\ref{sec:motivation}, we measure correspondence across H-H, R-R, and H-R image pairs using EPE and cosine similarity.
Results are shown in Table~\ref{tab:data_efficiency}.

Despite using significantly fewer training images, \modelname-DINO consistently outperforms the DINO baseline across metrics.
Even with a single manually labeled frame, \modelname-DINO achieves an EPE of 23.81, compared to 87.04 for DINO.
This demonstrates how our alignment objective is robust to limited supervision, enabling practical use cases where full forward kinematics are unavailable.
\begin{figure}[b]
    \centering
    \includegraphics[width=\linewidth]{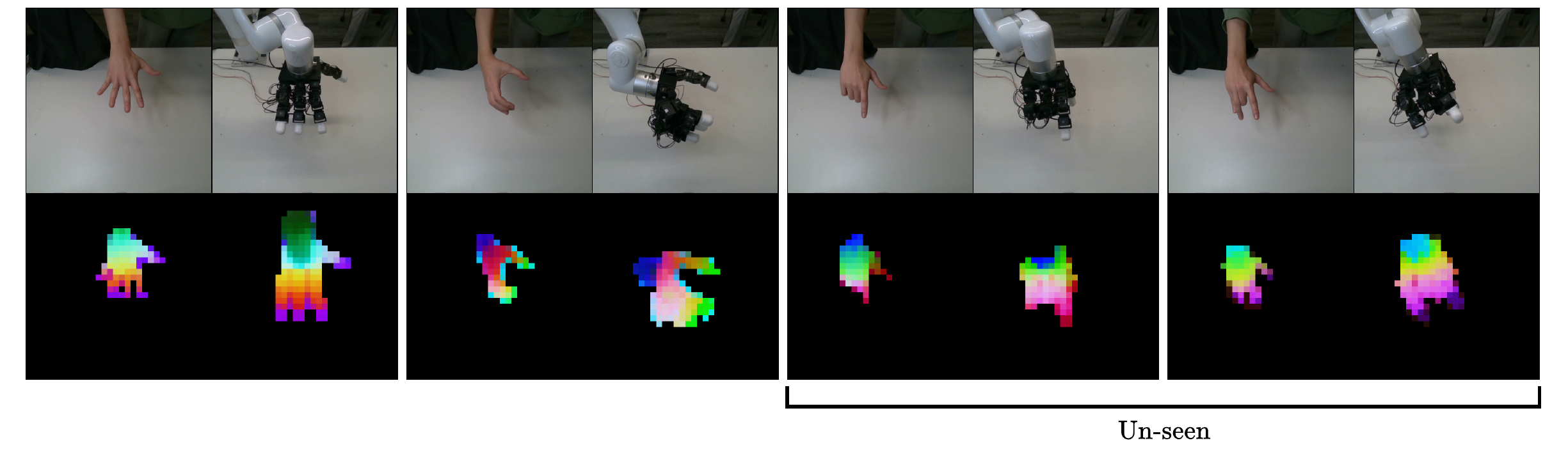}
    \caption{\textbf{Semantic alignment in diverse poses.} PCA is computed jointly on human and robot hand features. Semantic correspondence is strong even for poses unseen during training.}
    \label{fig:pose_pca}
\end{figure}
\subsection{Generalization to Unseen Poses}
\label{sec:pose_generalization}

Semantic features are pose-invariant---a thumb is a thumb regardless of its pose.
This allows training on unaligned pairs, and more importantly, enables generalization to poses never seen during training.
To verify this, we collect human-robot image pairs in diverse hand poses, some not seen during training.
We extract features from both embodiments and compute PCA jointly.
As shown in Figure~\ref{fig:pose_pca}, semantically corresponding parts map to a similar color across human and robot hands---even for poses unseen during training.

\begin{figure}[t]
    \centering
    \includegraphics[width=0.85\linewidth]{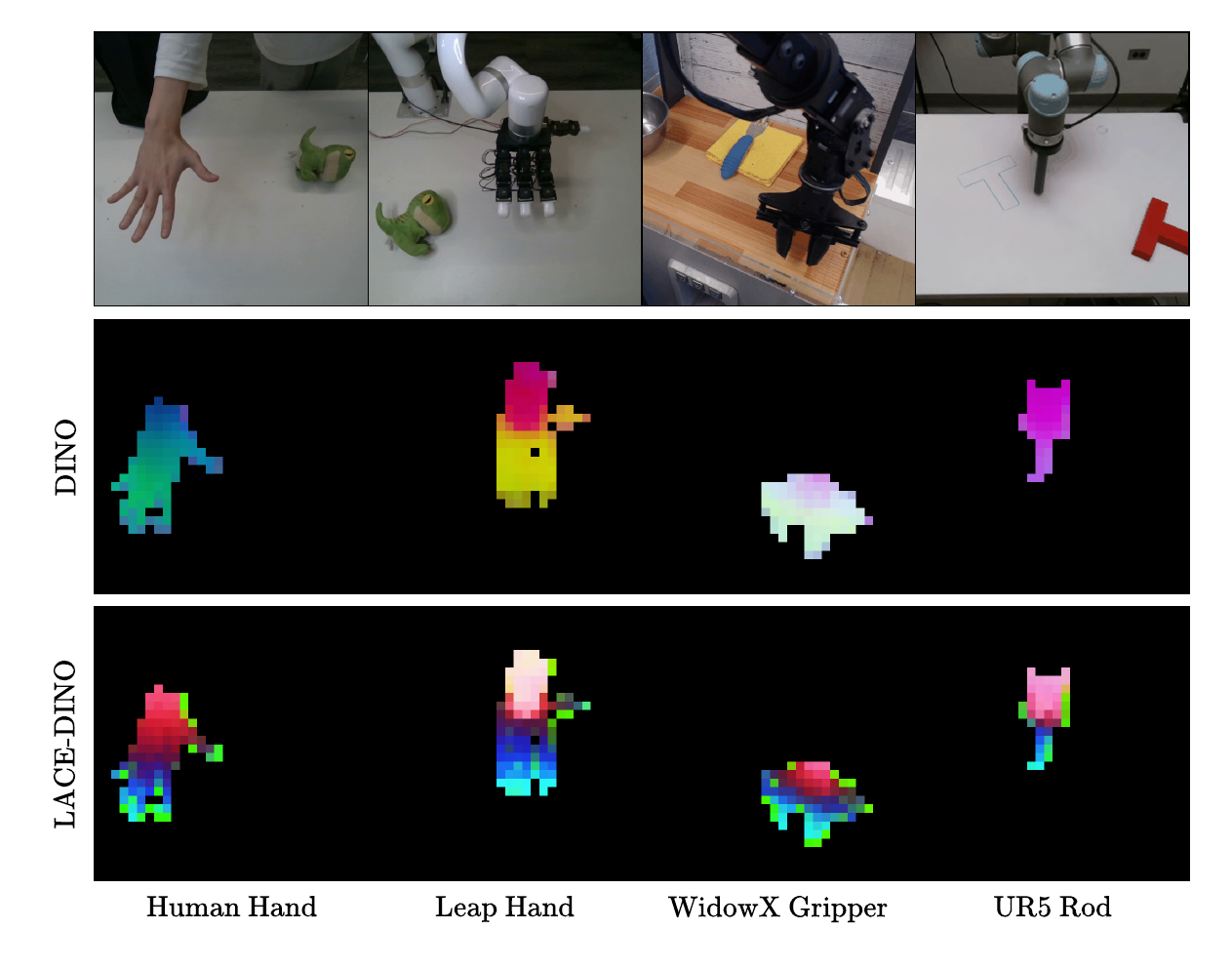}
    \caption{\textbf{Multi-embodiment alignment.} From left to right: human, Leap hand, WidowX Gripper, and UR5 Pole. We jointly align them using \modelname~and compute PCA on features of the robot patches. \modelname~can align multiple embodiments simultaneously.}
    \label{fig:multi_emb}
\end{figure}

\subsection{Generalization to Multiple Embodiments}
\label{sec:multi_emb}

Although our main experiments focus on human and dexterous robot hand alignment, \modelname{} is not limited to this setting: it can align embodiments of different morphologies and handle multiple embodiments at once.
We demonstrate this by jointly aligning four distinct embodiment types: human hand, multi-finger robot hand, a parallel-jaw gripper, and a rod end-effector.
We select one episode each from two Open X-Embodiment~\cite{openx2024} subsets: \texttt{bridge} for the WidowX gripper and \texttt{columbia\_cairlab\_pusht\_real} for the UR5 rod end-effector.
We manually annotate 10 random frames from each episode with keypoint correspondences.
Note that keypoint annotation is flexible and allow many-to-one or drop keyopint correspondence.

Following the augmentation protocol in Section~\ref{sec:implementation}, we augment each embodiment's dataset to 1K images.
Then, we train \modelname-DINO on all four embodiments jointly and compute PCA on the combined patch features of the robots.
As shown in Figure~\ref{fig:multi_emb}, semantically corresponding regions across all embodiments map to similar colors, demonstrating that \modelname~extends to multiple embodiments alignment despite a significant morphological difference.

\subsection{Preserving Feature Quality with Gram Loss}
\label{sec:gram_ablation}

Fine-tuning on limited data risks catastrophic forgetting, where the model overfits to the alignment task and loses its general-purpose representations.
The Gram loss acts as a stabilizer by preserving the feature semantic structure during training.
To demonstrate this, we train \modelname-DINO with varying Gram loss weights $\lambda$ and evaluate the last checkpoint on standard downstream tasks following the evaluation protocol in~\cite{simeoni2025dinov3,oquab2023dinov2}: linear classification on ImageNet using the CLS token and semantic segmentation on ADE20K using a linear head over dense features.
We also report human-to-robot alignment metrics to examine the trade-off between preservation and alignment.
Results are shown in Table~\ref{tab:gram_ablation}.

\vspace{2em}

Without Gram loss ($\lambda = 0$), downstream performance degrades significantly, confirming catastrophic forgetting.
Conversely, when $\lambda$ is too large, the model overly penalizes deviation from the original feature structure, limiting semantic alignment between embodiments.
At moderate values in between, \modelname-DINO achieves strong cross-embodiment alignment while maintaining downstream task performance comparable to pretrained DINO.

\begin{table}[t]
    \centering
    \caption{\textbf{Ablation on Gram loss weight $\lambda$.} We report downstream task performance (ImageNet classification accuracy, ADE20K segmentation mIoU) alongside alignment metrics. Moderate $\lambda$ balances feature preservation and alignment.}
    \vspace{0.5em}
    \label{tab:gram_ablation}
    \def\arraystretch{1.1}  
    \setlength\tabcolsep{0.6em}  
    \scalebox{0.85}{
    \begin{tabular}{lcccccccc}
        \toprule
        & {ImageNet} & {ADE20K} & \multicolumn{2}{c}{H-R} & \multicolumn{2}{c}{H-H} & \multicolumn{2}{c}{R-R} \\
        \cmidrule(lr){2-2} \cmidrule(lr){3-3} \cmidrule(lr){4-5} \cmidrule(lr){6-7} \cmidrule(lr){8-9}
        & Acc $\uparrow$ & mIoU $\uparrow$ & EPE $\downarrow$ & Cos $\uparrow$ & EPE $\downarrow$ & Cos $\uparrow$ & EPE $\downarrow$ & Cos $\uparrow$ \\
        \midrule
        DINO         & 76.73 & 42.12 & 87.04 & 0.473 & 34.94 & 0.679 & 23.00 & 0.824 \\
        \midrule
        \modelname-DINO \\
        \quad $\lambda=0$    & 73.10 & 38.49 & 19.17 & 0.921 & 23.03 & 0.817 & 19.73 & 0.886 \\
        \quad $\lambda=0.01$ & 74.17 & 40.44 & 19.37 & 0.927 & 18.63 & 0.869 & 17.27 & 0.921 \\
        \quad $\lambda=0.1$  & 74.42 & 40.58 & 19.21 & 0.926 & 17.78 & 0.903 & 16.04 & 0.948 \\
        \quad $\lambda=1$    & 74.86 & 40.41 & 19.57 & 0.910 & 17.33 & 0.919 & 15.99 & 0.958 \\
        \quad $\lambda=10$   & 75.12 & 40.54 & 20.15 & 0.862 & 17.14 & 0.921 & 16.42 & 0.959 \\
        \quad $\lambda=100$  & 75.05 & 40.79 & 24.33 & 0.790 & 17.34 & 0.920 & 16.62 & 0.952 \\
        \bottomrule
    \end{tabular}
    }
\end{table}

\subsection{Ablation on \modelname~Components}
\label{sec:component_ablation}

We ablate over the design choices in \modelname-DINO to understand their individual contributions.
Table~\ref{tab:component_ablation} reports alignment metrics across all embodiment pairs (Human-Robot, Human-Human, Robot-Robot) following the protocol in Section~\ref{sec:motivation}, along with downstream task performance (ImageNet, ADE20k) following~\ref{sec:gram_ablation}.
The default setting achieves strong alignment while preserving general feature quality, as reflected in downstream task performance. Below, we further describe each component in detail.

\begin{table}[t]
    \centering
    \caption{\textbf{Ablation on \modelname-DINO components.} Each row removes or modifies one component from the default setting.}
    \vspace{0.5em}
    \label{tab:component_ablation}
    \scalebox{0.9}{
    \begin{tabular}{lcccccccc}
        \toprule
        & {ImageNet} & {ADE20k} & \multicolumn{2}{c}{H-R} & \multicolumn{2}{c}{H-H} & \multicolumn{2}{c}{R-R} \\
        \cmidrule(lr){2-2} \cmidrule(lr){3-3} \cmidrule(lr){4-5} \cmidrule(lr){6-7} \cmidrule(lr){8-9}
        & Acc $\uparrow$ & mIoU $\uparrow$ & EPE $\downarrow$ & Cos $\uparrow$ & EPE $\downarrow$ & Cos $\uparrow$ & EPE $\downarrow$ & Cos $\uparrow$ \\
        \midrule
        \modelname-DINO               & \textbf{74.86} & \textbf{40.41} & 19.36 & 0.910 & 17.57 & 0.900 & 16.00 & 0.945 \\
        \quad w/o EMA           & 74.01 & 37.16 & 19.04 & 0.872 & 17.50 & 0.872 & 14.89 & 0.925 \\
        \quad w/o intra-class   & 74.28 & 40.06 & 18.91 & 0.876 & 18.74 & 0.872 & 15.17 & 0.942 \\
        \quad $\rightarrow$ uni-dir & 73.39 & 38.95 & 19.51 & 0.880 & 17.94 & 0.911 & 15.77 & 0.944 \\
        \quad $\rightarrow$ fwd KL & 73.34 & 38.69 & 18.91 & 0.907 & 17.29 & 0.904 & 15.51 & 0.946 \\
        \bottomrule
    \end{tabular}
    }
\end{table}

\paragraph{EMA for target features.}
The semantic structure of robot embodiments in pretrained DINO is usually determined by surface features such as texture or color (e.g., white fingertips).
Using an exponential moving average for the target allows the semantic structure to gradually evolve toward structural alignment based on corresponding body parts.

\paragraph{Intra-class pairs.}
We include same-embodiment pairs (human-human, robot-robot) alongside cross-embodiment pairs.
We expect this to encourage granular semantic alignment—differentiating sub-parts (thumb, index, etc.) rather than treating the hand as a single semantic unit.

\paragraph{Bidirectional alignment.}
We use bidirectional alignment (both human-to-robot and robot-to-human) rather than uni-directional alignment (robot-to-human only).
We expect this to produce better correspondence by allowing both embodiments to adjust toward each other.

\paragraph{Reverse KL divergence.}
We use reverse KL rather than forward KL for the alignment loss.
Forward KL aggressively matches the source distribution to the target, which can lead to unstable training; reverse KL provides a softer alignment signal.

\section{Conclusion and Discussion}
\label{sec:conclusion}

We identified that the visual gap between human and robot embodiments propagates into self-supervised visual representations.
To address this, we presented \modelname, a fine-tuning framework that leverages the backbone's inherent semantic priors to align embodiment representations in latent space.
Using sparse keypoint supervision from a single robot demonstration, \modelname{} lifts correspondence to dense features while preserves general visual understanding.
Our experiments validate that this latent-space alignment benefits policy learning in low robot data regimes and enables generalization to unseen environments—without additional inference cost.

\section*{Acknowledgements}
This research was financially supported by the Ministry of Trade, Industry, and Energy (MOTIE), Korea, under the ``Global Industrial Technology Cooperation Center (GITCC) program'' supervised by the Korea Institute for Advancement of Technology (KIAT) (Task No. P0028420).
This research was supported by the National Research Council of Science \& Technology (NST) grant by the Korea government (MSIT) (No. GTL25041-000).

\bibliographystyle{unsrtnat}
\bibliography{main}

\appendix
\appendix

\begin{center}
{\huge Appendix}
\end{center}

\section{Localization Accuracy for Policy}
\label{app:localization}

In the main paper, we report policy success rates for object grasping, but failures can arise from both localization errors and motion primitive variability.
To isolate the contribution of visual representation, we evaluate localization accuracy directly without executing motions.

We train a DETR-based policy~\ref{sec:policy} to predict the embodiment center and target object center, and report \textit{hit rate}: whether the predicted point falls within the ground-truth bounding box.
We train three \modelname-DINO variants on different environments (Lab, Kitchen, Office), while the downstream policy is always trained and evaluated in the Lab environment.
Table~\ref{tab:localization} reports results across data regimes based on percentage of robot data used: zero (0\%), few (10\%), and co (100\%), and evaluation settings: In-domain and OOD-env (human data only).

\begin{table}[ht]
    \centering
    \caption{\textbf{Controlled policy evaluation.} We evaluate localization accuracy directly, independent of motion execution. We report hit rate (\%) for embodiment (Emb), object (Obj), and both, as well as L1 distance between predicted and ground-truth points. We evaluate policies across data regimes (zero, few, co) and scene settings (in-domain, OOD-env). Backbone suffixes (e.g., Kitchen, Office) indicate the representation training environment, demonstrating that the representation generalizes to unseen environments.}
    \label{tab:localization}
    \def\arraystretch{1.1}
    \setlength\tabcolsep{0.4em}
    \scalebox{0.8}{
    \begin{tabular}{llcccccccc}
        \toprule
        & & \multicolumn{4}{c}{In-domain} & \multicolumn{4}{c}{OOD-env} \\
        \cmidrule(lr){3-6} \cmidrule(lr){7-10}
        Regime & Vision Backbone & Emb $\uparrow$ & Obj $\uparrow$ & Both $\uparrow$ & L1 $\downarrow$ & Emb $\uparrow$ & Obj $\uparrow$ & Both $\uparrow$ & L1 $\downarrow$ \\
        \midrule
        zero & DINO & 21.1 & 71.1 & 16.3 & 0.130 & 0.0 & 54.5 & 0.0 & 0.149 \\
        \cmidrule(lr){2-10}
             & \modelname-DINO (Lab) & 78.4 & 93.7 & 74.7 & 0.063 & 59.1 & 94.5 & 59.1 & 0.069 \\
             & \modelname-DINO (Kitchen) & 82.6 & 86.8 & 74.2 & 0.060 & 74.5 & 91.8 & 71.8 & 0.073 \\
             & \modelname-DINO (Office) & 93.2 & 92.1 & 85.8 & 0.044 & 64.5 & 94.5 & 59.1 & 0.055 \\
        \midrule
        few  & DINO & 100.0 & 83.2 & 83.2 & 0.051 & 94.5 & 69.1 & 68.2 & 0.054 \\
        \cmidrule(lr){2-10}
             & \modelname-DINO (Lab) & 100.0 & 94.2 & 94.2 & 0.030 & 88.2 & 96.4 & 84.5 & 0.032 \\
             & \modelname-DINO (Kitchen) & 100.0 & 91.6 & 91.6 & 0.031 & 80.9 & 97.3 & 78.2 & 0.042 \\
             & \modelname-DINO (Office) & 100.0 & 94.7 & 94.7 & 0.028 & 99.1 & 99.1 & 98.2 & 0.030 \\
        \midrule
        co   & DINO & 100.0 & 92.6 & 92.6 & 0.032 & 100.0 & 90.9 & 90.9 & 0.025 \\
        \cmidrule(lr){2-10}
             & \modelname-DINO (Lab) & 100.0 & 94.2 & 94.2 & 0.026 & 100.0 & 95.5 & 95.5 & 0.024 \\
             & \modelname-DINO (Kitchen) & 100.0 & 94.7 & 94.7 & 0.024 & 100.0 & 100.0 & 100.0 & 0.019 \\
             & \modelname-DINO (Office) & 100.0 & 94.7 & 94.7 & 0.023 & 100.0 & 99.1 & 99.1 & 0.019 \\
        \bottomrule
    \end{tabular}
}
\end{table}

\paragraph{Recovering robot state in zero-shot.}
In the zero-shot regime, \modelname-DINO successfully recovers robot state from visual observations without any robot training data.
DINO fails at embodiment localization (21.1\% In-domain, 0\% OOD-env), while \modelname-DINO (Lab) achieves 78.4\% and 59.1\% respectively.

\paragraph{Skill transfer.}
In low data regimes and OOD-env setting, in which human data dominates, \modelname-DINO consistently outperforms DINO on object hit rate (Obj), indicating that object localization transfer from human to robot observations. Here, transfer in object localization is analogous to skill transfer.

\paragraph{Environment robustness.}
\modelname-DINO (Kitchen) and \modelname-DINO (Office) perform comparably to \modelname-DINO (Lab) despite the environment mismatch in representation learning and downstream task, confirming that \modelname generalizes across environments.

\paragraph{Localization precision.}
We also report L1 distance to measure localization precision.
\modelname-DINO achieves substantially lower L1 distance than DINO.
Lower localization error makes motion primitives less sensitive during execution, which explains the success rate gap in real-world evaluation.

\begin{table}[t]
    \centering
    \caption{\textbf{Keypoint alignment across backbones.} Following \cref{sec:motivation}, we compare human--robot alignment with and without \modelname~applied across different SSL backbones.
    \modelname~consistently improves alignment regardless of the backbone choice.}
    \label{tab:backbone_alignment}
    \def\arraystretch{1.1}
    \setlength\tabcolsep{0.5em}
    \scalebox{0.85}{
    \begin{tabular}{llcccc}
        \toprule
        & & \multicolumn{2}{c}{EPE $\downarrow$} & \multicolumn{2}{c}{Cos $\uparrow$} \\
        \cmidrule(lr){3-4} \cmidrule(lr){5-6}
        Setting & Backbone & Base & Ours & Base & Ours \\
        \midrule
        Human--Robot & DINO & 87.04 & 19.36 & 0.473 & 0.910 \\
                     & SigLIP~\cite{tschannen2025siglip} & 107.16 & 28.77 & 0.085 & 0.436 \\
                     & I-JEPA~\cite{assran2023self} & 81.42 & 20.51 & 0.242 & 0.630 \\
                     & V-JEPA2~\cite{assran2025v} & 83.51 & 30.32 & 0.387 & 0.637 \\
        \bottomrule
    \end{tabular}
}
\end{table}

\section{Other SSL Backbones}
\label{sec:other_ssl}

The main paper demonstrates \modelname~using DINO, chosen for its popularity in robot learning and strong semantic feature encoding.
Here we verify that \modelname~generalizes to other self-supervised backbones by evaluating alignment metrics on SigLIP~\cite{tschannen2025siglip}, I-JEPA~\cite{assran2023self}, and V-JEPA2~\cite{assran2025v}.

We follow the same experimental protocol as Section~\ref{sec:motivation}:
we train on keypoint-annotated images from a subset of HInt for human hand, and random frames from a single LeapHand demonstration for robot hand, then evaluate correspondence on a held-out test set.
We report two metrics: \textit{End Point Error (EPE)}, which finds the most similar patch in the target image via cosine similarity and measures pixel distance from its center to the ground-truth keypoint, and \textit{Cosine Similarity (Cos)}, which measures feature similarity between corresponding keypoint patches.
Lower EPE and higher Cos indicate better alignment.

Table~\ref{tab:backbone_alignment} shows results across all four backbones.
All models show substantial improvement in human--robot correspondence after applying \modelname, confirming that our framework is not specific to DINO.

\begin{table}[t]
    \centering
    \caption{\textbf{Zero-shot policy evaluation across backbones.} To evaluate whether \modelname~benefits cross-embodiment learning, we compare policies using backbones with and without \modelname~applied, following \cref{app:localization}. We report hit rate (\%) for embodiment (Emb), object (Obj), and both. All models are evaluated in the zero-shot regime (0\% robot data) on in-domain environment.}
    \label{tab:localization_zeroshot}
    \def\arraystretch{1.1}
    \setlength\tabcolsep{0.4em}
    \scalebox{0.85}{
    \begin{tabular}{llcccccc}
        \toprule
        & & \multicolumn{6}{c}{In-domain} \\
        \cmidrule(lr){3-8}
        & & \multicolumn{2}{c}{Emb $\uparrow$} & \multicolumn{2}{c}{Obj $\uparrow$} & \multicolumn{2}{c}{Both $\uparrow$} \\
        \cmidrule(lr){3-4} \cmidrule(lr){5-6} \cmidrule(lr){7-8}
        Regime & Backbone & Base & Ours & Base & Ours & Base & Ours \\
        \midrule
        zero & DINO & 21.1 & 78.4 & 71.1 & 93.7 & 16.3 & 74.7 \\
             & SigLIP~\cite{tschannen2025siglip} & 22.6 & 94.2 & 61.1 & 66.3 & 16.8 & 61.6 \\
             & I-JEPA~\cite{assran2023self} & 13.7 & 72.1 & 38.9 & 72.6 & 6.8 & 52.6 \\
             & V-JEPA2~\cite{assran2025v} & 3.7 & 56.8 & 88.9 & 91.6 & 3.2 & 54.2 \\
        \bottomrule
    \end{tabular}
}
\end{table}

We further verify that improved alignment translates to better cross-embodiment policy transfer.
Following the evaluation protocol in \cref{app:localization}, we train policies using pretrained and \modelname~backbones and evaluate in the zero-shot regime (0\% robot data) on the in-domain environment.
Table~\ref{tab:localization_zeroshot} reports hit rates for embodiment (Emb), object (Obj), and joint localization (Both).
Across all backbones, pretrained models achieve low joint hit rates (3.2--16.8\%), while \modelname~variants reach 52.6--74.7\%.
This consistent improvement confirms that the alignment gains observed in Table~\ref{tab:backbone_alignment} directly benefit downstream policy learning.

We note that \modelname~relies on backbones that encode semantic structure in their patch features.
Reconstruction-based models such as MAE~\cite{he2022masked}, which prioritize pixel-level rather than semantic representations, are less suited to this approach.

\section{Selective Feature Drift with Gram Loss}
\label{sec:feature_drift}

Feature quality and feature drift are distinct: a rotated feature space preserves relative structure (quality intact) yet shifts every vector (severe drift).
Low drift matters because VLAs and VLMs pretrained on vision backbone features~\cite{team2024octo} require expensive retraining if the entire feature space shifts, though far less if drift is confined to specific categories. 
Our goal is \emph{selective drift}---hand features should move to achieve alignment, while general object features remain stable for plug-and-play integration.

To visualize drift, we extract patch features from the same image using both pretrained DINO and \modelname-DINO, mask out background patches, and compute PCA on the combined set. Projecting both into a shared space reveals how far \modelname-DINO features have shifted from their DINO counterparts.

\begin{figure}[t]
\centering
\includegraphics[width=0.7\linewidth]{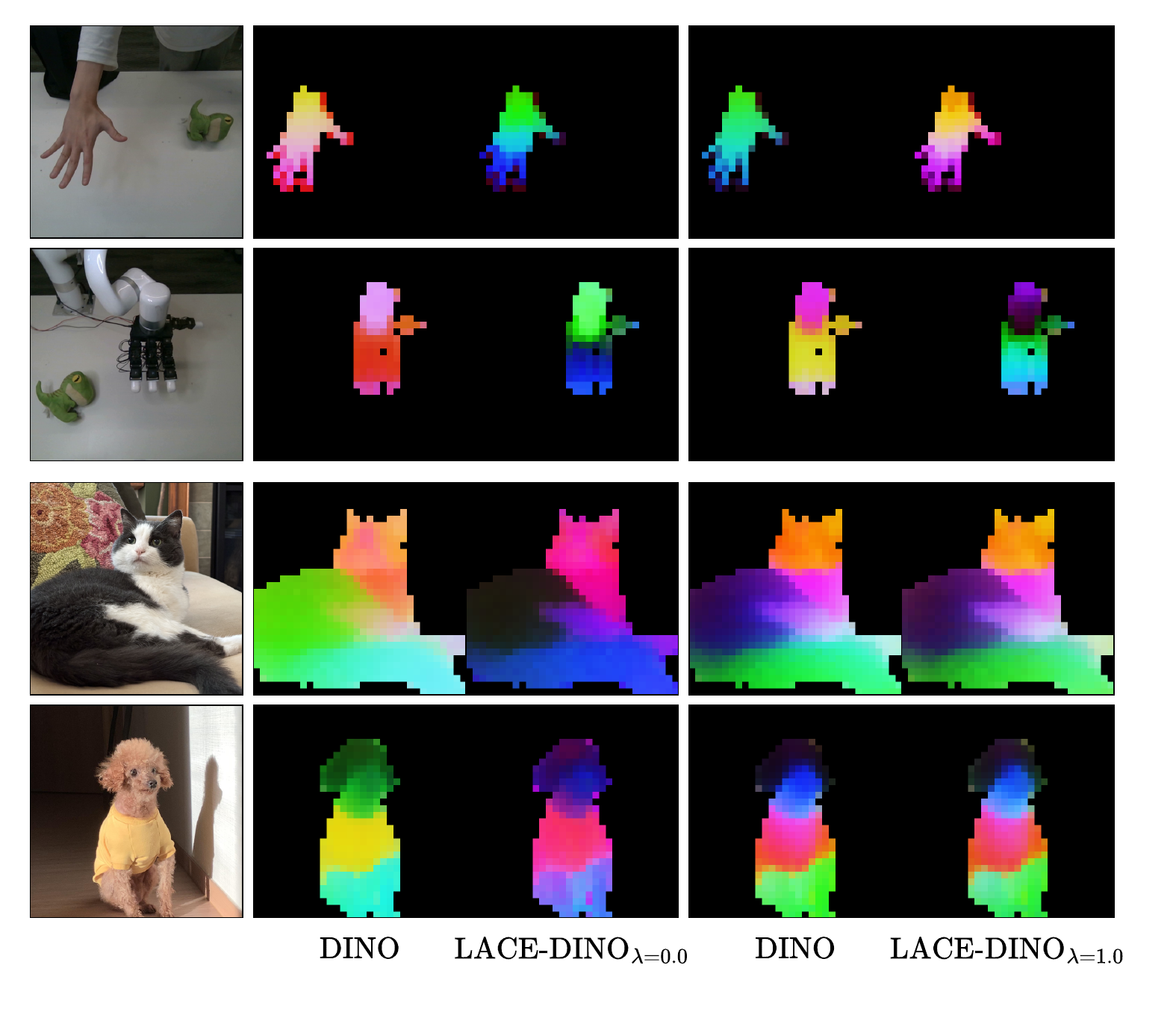}
\caption{\textbf{Feature drift from DINO.} PCA visualization of DINO and \modelname-DINO patch features extracted from the same image. Differing colors indicate drift. With Gram loss ($\lambda=1$), drift is confined to hand regions. (DINO's appearance varies because of different projection basis.)}
\label{fig:drift}
\end{figure}

Figure~\ref{fig:drift} shows results for our default setting ($\lambda=1$) and without Gram loss ($\lambda=0$). With Gram loss, drift is selective: hand features shift as intended, while object features remain consistent with pretrained DINO. Without Gram loss, drift is uncontrolled---hands and objects all shift indiscriminately. This confirms Gram loss is essential for confining representation changes to supervised categories. 

\section{Object Generalization}
\label{app:object_generalization}

Alignment fine-tuning may risk degrading DINO's semantic generalization.
We conduct a small-scale study to examine this, using visually distinct but semantically equivalent target objects.
Due to its rich semantic features, DINO is expected to generalize to novel object appearances within the same semantic class; we aim to show that \modelname-DINO retains this capability.

\begin{figure}[b]
    \begin{minipage}{0.5\linewidth}
        \centering
        \includegraphics[width=0.8\linewidth]{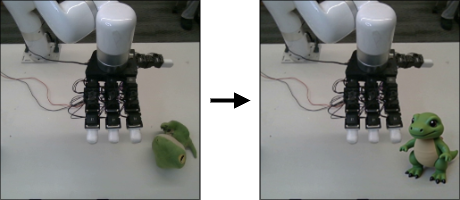}
        \captionof{figure}{\textbf{OOD-obj examples.} Using FLUX.1 Kontext, we edit the original target object (green dinosaur doll) into visually different variants of the same semantic class. These edited images are used to evaluate whether semantic generalization is preserved after \modelname.}
        \label{fig:obj_gen}
    \end{minipage}
    \hfill
    \begin{minipage}{0.4\linewidth}
        \centering
        \def\arraystretch{1.1}
        \setlength\tabcolsep{0.5em}
        \scalebox{0.85}{
        \begin{tabular}{lcc}
            \toprule
            & In-domain & OOD-obj \\
            \cmidrule(lr){2-2} \cmidrule(lr){3-3}
            & Obj $\uparrow$ & Obj $\uparrow$ \\
            \midrule
            DINO & 84.7 & 84.2 \\
            \modelname-DINO & 88.9 & 86.3 \\
            \bottomrule
        \end{tabular}
        }
        \captionof{table}{\textbf{Object generalization.} Obj hit rate (\%) for localizing the target object. In-domain uses the original object; OOD-obj uses visually edited variants of the same semantic class. \modelname-DINO achieves comparable results, indicating object generalization is retained.}
        \label{tab:obj_gen}
    \end{minipage}
\end{figure}

Using FLUX.1 Kontext~\cite{batifol2025flux}, we edit test images to replace the original target object (green dinosaur doll) with visually different variants of the same semantic class; we term this setting OOD-obj (Figure~\ref{fig:obj_gen}).

Following Section~\ref{sec:policy}, we train a DETR-based policy for object localization using robot data only.
As shown in Table~\ref{tab:obj_gen}, both methods achieve similar Obj hit rates on OOD-obj, indicating that alignment fine-tuning does not degrade DINO's semantic generalization to novel object appearances.

\section{Implementation Details}
\label{app:implementation}

Implementation details are described in the main paper (Sections~\ref{sec:implementation}--\ref{sec:policy}). Here we provide additional hyperparameters.

\subsection{\modelname-DINO Training}

We full fine-tune DINOv3 ViT-S/16 pretrained on LVD-1689M.
We use a batch size of 64 with AdamW optimizer (weight decay 0.05, $\beta = (0.9, 0.999)$, max gradient norm 1.0) and train for 10K iterations with seed 42.
Training is performed on a single NVIDIA RTX 3090 (24GB VRAM) and takes approximately 1 hour.

\subsection{Linear Probing}

We freeze the \modelname-DINO backbone and extract features from the last layer.
We add sinusoidal 2D position embeddings to the patch features, apply attention pooling to aggregate over patches, and use a linear head for prediction.
We use a batch size of 32 with AdamW optimizer (learning rate 1e-3, weight decay 0.05, $\beta = (0.9, 0.999)$, max gradient norm 1.0).
We apply linear warmup for 10\% of training followed by cosine decay.
We use MSE loss and train for 10K iterations with seed 0.
Training takes approximately 2 hours on the same hardware.

\subsection{DETR Policy}

We use a DETR-based policy with frozen \modelname-DINO as the vision encoder.
The transformer encoder and decoder are initialized from pretrained DETR weights.
The prediction head outputs two 2D points: end-effector position and target object position.
We use separate learning rates: 1e-5 for pretrained DETR parameters and 1e-4 for new parameters.
We apply linear warmup for 10\% of training followed by cosine decay.
All downstream experiments use seed 0.
Training takes approximately 3 hours on the same hardware.

\begin{figure}[ht]
    \centering
    \includegraphics[width=\linewidth]{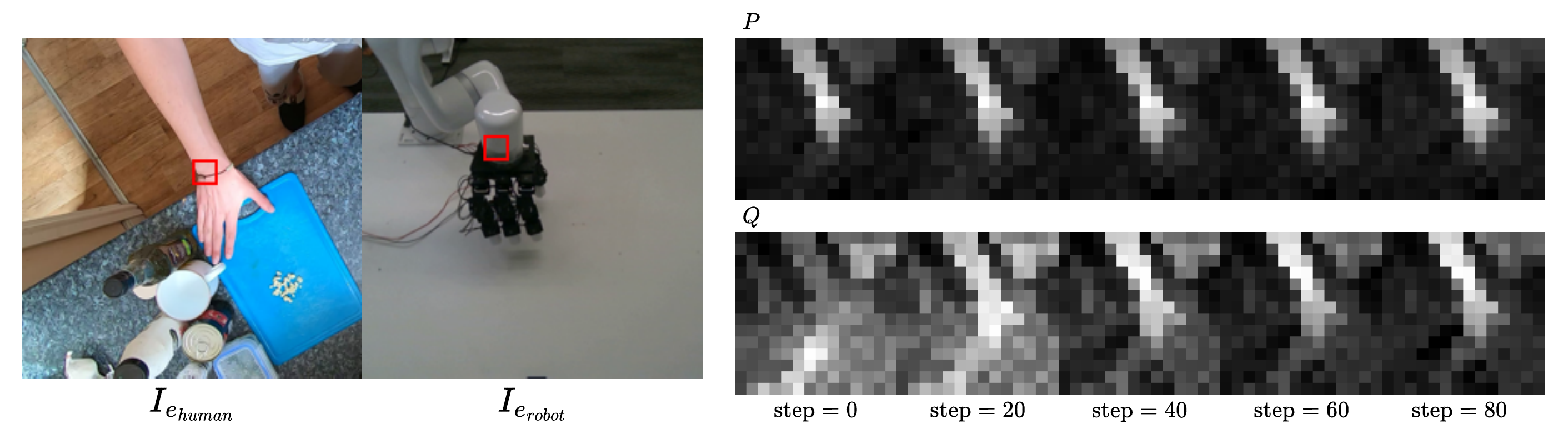}
    \caption{\textbf{Evolution of similarity distributions during training.} We visualize the self-similarity distribution $P$ (top) and cross-similarity distribution $Q$ (bottom) as defined in Section~\ref{sec:semantic_alignment_loss}. Red-bordered squares indicate corresponding patches. $P$ captures how a human keypoint patch relates to all patches within the human image, while $Q$ captures how the corresponding robot keypoint patch relates to the same human patches. As training progresses, $Q$ converges toward $P$ via reverse KL divergence. Distribution values are min-max normalized for better visualization.}
    \label{fig:loss_viz}
\end{figure}

\begin{algorithm}[t]
    \caption{\modelname~Training Step}
    \label{alg:lace_torch}
    \definecolor{codegreen}{rgb}{0.0, 0.5, 0.5}
    \newcommand{\cmt}[1]{{\color{codegreen}\# #1}}
    \begin{algorithmic}[0]
        \ttfamily
        \STATE \cmt{$\tilde{f}_\theta$: pretrained backbone (L2-normalized), $\tilde{f}_{\bar{\theta}}$: EMA model}
        \STATE \cmt{$\tau$: temperature, $\lambda$: Gram weight, $\alpha$: EMA momentum}
        \STATE
        \STATE \cmt{$\mathcal{K}$: shared keypoints, $p_a$, $p_b$: patch indices for all $k \in \mathcal{K}$}
        \STATE $I_a$, $I_b$, $p_a$, $p_b$ = sample\_batch()
        \STATE
        \STATE \cmt{feature extraction}
        \STATE $Z_a$, $Z_b$ = $\tilde{f}_\theta(I_a)$, $\tilde{f}_\theta(I_b)$ \cmt{[B, N, C]}
        \STATE \textbf{with} no\_grad():
        \STATE \quad $\bar{Z}_a$ = $\tilde{f}_{\bar{\theta}}(I_a)$
        \STATE
        \STATE \cmt{alignment loss}
        \STATE \textbf{for} $k \in \mathcal{K}$ \textbf{do}
        \STATE \quad $Q_k$ = softmax($Z_b[p_b^k]$ @ $Z_a$.T / $\tau$) \cmt{cross-similarity}
        \STATE \quad $P_k$ = softmax($\bar{Z}_a[p_a^k]$ @ $\bar{Z}_a$.T / $\tau$) \cmt{self-similarity}
        \STATE $\mathcal{L}_{\text{align}}$ = $\frac{1}{|\mathcal{K}|}$ $\sum_{k}$ $D_{\text{KL}}$($Q_k$ $\|$ $P_k$) \cmt{reverse KL}
        \STATE
        \STATE \cmt{Gram loss}
        \STATE $\mathcal{L}_{\text{gram}}$ = MSE($Z_a$ @ $Z_a$.T, $\bar{Z}_a$ @ $\bar{Z}_a$.T)
        \STATE
        \STATE \cmt{update}
        \STATE loss = $\mathcal{L}_{\text{align}}$ + $\lambda$ * $\mathcal{L}_{\text{gram}}$
        \STATE loss.backward()
        \STATE optimizer.step()
        \STATE $\bar{\theta}$ = $\alpha$ * $\bar{\theta}$ + (1 - $\alpha$) * $\theta$ \cmt{EMA update}
    \end{algorithmic}
\end{algorithm}

\end{document}